\documentclass{article}

\usepackage[final]{ewrl_2025}

\usepackage[utf8]{inputenc} 
\usepackage[T1]{fontenc}    
\usepackage{hyperref}       
\usepackage{url}            
\usepackage{booktabs}       
\usepackage{amsfonts}       
\usepackage{nicefrac}       
\usepackage{microtype}      
\usepackage{xcolor}         

\usepackage{mathtools}
\usepackage{amsthm}
\theoremstyle{plain}
\newtheorem{theorem}{Theorem}[section]
\theoremstyle{definition}
\newtheorem{definition}[theorem]{Definition}

\usepackage{algorithm}
\usepackage{algorithmic}

\title{Off-Policy Maximum Entropy RL with Future State and Action Visitation Measures}

\author{%
  Adrien Bolland\\
  Montefiore Institute\\
  University of Liège\\
  Liège, Belgium\\
  \texttt{adrien.bolland@uliege.be} \\
  \And
  Gaspard Lambrechts\\
  Montefiore Institute\\
  University of Liège\\
  Liège, Belgium\\
  \texttt{gaspard.lambrechts@uliege.be} \\
  \And
  Damien Ernst\\
  Montefiore Institute\\
  University of Liège\\
  Liège, Belgium\\
  \texttt{dernst@uliege.be} \\
}

\begin{document}

\maketitle

\begin{abstract}
Maximum entropy reinforcement learning integrates exploration into policy learning by providing additional intrinsic rewards proportional to the entropy of some distribution. In this paper, we propose a novel approach in which the intrinsic reward function is the relative entropy of the discounted distribution of states and actions (or features derived from these states and actions) visited during future time steps. This approach is motivated by two results. First, a policy maximizing the expected discounted sum of intrinsic rewards also maximizes a lower bound on the state-action value function of the decision process. Second, the distribution used in the intrinsic reward definition is the fixed point of a contraction operator. Existing algorithms can therefore be adapted to learn this fixed point off-policy and to compute the intrinsic rewards. We finally introduce an algorithm maximizing our new objective, and we show that resulting policies have good state-action space coverage and achieve high-performance control.
\end{abstract}

\section{Introduction} \label{sec:introduction}

Many challenging tasks where an agent makes sequential decisions have been solved with reinforcement learning (RL). Examples range from playing games \citep{mnih2015human, silver2017mastering}, or controlling robots \citep{kalashnikov2018qt, haarnoja2018composable}, to managing energy systems and markets \citep{boukas2021deep, aittahar2014optimal}. In practice, many RL algorithms are applied in combination with an exploration strategy to achieve high-performance control. Assuming the agent takes actions in a Markov decision process (MDP), these exploration strategies usually consist of providing intrinsic reward bonuses to the agent for achieving certain behaviors. Typically, the bonus enforces taking actions that reduce the uncertainty about the environment \citep{pathak2017curiosity, burda2018large, zhang2021noveld}, or actions that enhance the variety of states and actions in trajectories \citep{bellemare2016unifying, lee2019efficient, guo2021geometric, williams1991function, haarnoja2019soft}. In many of the latter methods, the intrinsic reward function is the entropy of some distribution over the state-action space. Optimizing jointly the reward function of the MDP and the intrinsic reward function, in order to eventually obtain a high-performing policy, is called Maximum Entropy RL (MaxEntRL) and was shown to be effective in many problems.

The reward of the MDP was already extended with the entropy of the policy in early algorithms \citep{williams1991function} and was only later called MaxEntRL \citep{ziebart2008maximum, toussaint2009robot}. This particular reward regularization provides substantial improvements in the robustness of the resulting policy \citep{ziebart2010modeling, husain2021regularized, brekelmans2022your} and provides a learning objective function with good smoothness and concavity properties \citep{ahmed2019understanding, bolland2023policy}. Several commonly used algorithms can be named, like soft Q-learning \citep{haarnoja2017reinforcement, schulman2017equivalence} and soft actor-critic \citep{haarnoja2018soft, haarnoja2019soft}. This MaxEntRL objective nevertheless only rewards the randomness of actions and neglects the influence of the policy on the visited states, which, in practice, may lead to inefficient exploration.

In order to enhance exploration, \citet{hazan2019provably} were the first to propose to intrinsically motivate agents to have a uniform discounted visitation measure over states. Several works have afterward been developed to maximize the entropy of the discounted state visitation measure and the stationary state visitation measure. For discrete state and action spaces, optimal exploration policies, which maximize the entropy of these visitation measures, can be computed to near optimality with off-policy tabular model-based RL algorithms \citep{hazan2019provably, mutti2020intrinsically, tiapkin2023fast}. For continuous state and action spaces, alternative methods rely on $k$ nearest neighbors to estimate the density of the visitation measure of states (or features built from the states) and compute the intrinsic rewards, which can afterward be optimized with any RL algorithm \citep{liu2021behavior, yarats2021reinforcement, seo2021state, mutti2021task}. These methods require sampling new trajectories at each iteration; they are on-policy, and estimating the intrinsic reward function is computationally expensive. Some other methods rely on parametric density estimators to reduce the computational complexity and share information across learning steps \citep{lee2019efficient, guo2021geometric, islam2019marginalized, zhang2021exploration}. The additional function approximator is typically learned on-policy by maximum likelihood estimation based on batches of truncated trajectories. Alternative methods have adapted this MaxEntRL objective to maximize entropy of states visited in single trajectories \citep{mutti2022importance, jain2024maximum}. When large and/or continuous state and action spaces are involved, relying on parametric function approximators is likely the best choice. Nevertheless, existing algorithms are on-policy. They require sampling new trajectories from the environment at (nearly) every update of the policy, and cannot be applied using a buffer of arbitrary transitions, in batch-mode RL, or in continuing tasks. Furthermore, learning the discounted visitation measure is more desirable than learning the stationary one, but may be challenging in practice due to the exponentially decreasing influence of the time step at which states are visited \citep{islam2019marginalized}.

The main contribution of this paper is to introduce a MaxEntRL objective relying on a new intrinsic reward function for exploring effectively the state and action spaces, which also alleviates the previous limitations. This intrinsic reward function is the relative entropy of the discounted distribution of states and actions (or features from these states and actions) visited during the next time steps. We prove two results motivating the MaxEntRL objective. First, a policy maximizing the expected discounted sum of intrinsic rewards also maximizes a lower bound on the state-action value function of the decision process. Second, the visitation distribution used in the new intrinsic reward function is the fixed point of a contraction operator. Existing RL algorithms can integrate an additional learning step to approximate this fixed point off-policy, using N-step state-action transitions and bootstrapping the operator. It is then possible to approximate the intrinsic reward function and learn a policy maximizing the extended rewards with the adapted algorithm. We illustrate this methodology on off-policy actor-critic \citep{degris2012off}. The resulting MaxEntRL algorithm is off-policy; it efficiently computes exploration policies with good discounted visitation probability coverage and high-performing control policies.

The visitation measure of future states and actions, which we use to extend the reward function in this article, has a well-established history in the development of RL algorithms. It was popularized by \citet{janner2020generative}, who learned the distribution of future states as a generalization of the successor features \citep{barreto2017successor}. They demonstrated that this distribution allows expressing the state-action value function by separating the influence of the dynamics and the reward function, and that it could be learned off-policy by exploiting its recursive expression. Several algorithms have been proposed to learn this distribution, either by maximum likelihood estimation \citep{janner2020generative}, by contrastive learning \citep{mazoure2023contrastive}, or using diffusion models \citep{mazoure2023value}. These distributions of future states and actions have found applications in goal-based RL \citep{eysenbach2020c, eysenbach2022contrastive}, in offline pre-training with expert examples \citep{mazoure2023accelerating}, in model-based RL \citep{ma2023learning}, or in planning \citep{eysenbach2023contrastive}. We are the first to integrate them into the MaxEntRL framework for enhancing exploration through learning.

The manuscript is organized as follows. In Section \ref{sec:prelim}, the RL problem is reminded, and the MaxEntRL framework is formulated. In Section \ref{sec:maxentrl_visitation}, we introduce and discuss a new MaxEntRL objective. Section \ref{sec:off_pol_algo} details how to learn a model of the conditional state visitation probability that allows estimating this new objective. We finally present experimental results in Section \ref{sec:experiments} and conclude in Section \ref{sec:conclusion}.

\section{Background and Preliminaries} \label{sec:prelim}

\subsection{Markov Decision Processes}
This paper focuses on problems in which an agent makes sequential decisions in a stochastic environment \citep{sutton2018reinforcement}. The environment is modeled with an infinite-time Markov decision process (MDP) composed of a state space $\mathcal{S}$, an action space $\mathcal{A}$, an initial state distribution $p_0$, a transition distribution $p$, a bounded reward function $R$, and a discount factor $\gamma \in [0, 1)$. Agents interact in this MDP by providing actions sampled from a policy $\pi$. During this interaction, an initial state $s_0 \sim p_0(\cdot)$ is first sampled, then, the agent provides at each time step $t$ an action $a_t \sim \pi(\cdot| s_t)$ leading to a new state $s_{t+1} \sim p(\cdot|s_t, a_t)$. In addition, after each action $a_t$ is executed, a reward $r_{t} = R(s_t, a_t) \in \mathbb{R}$ is observed. We denote the expected return of the policy $\pi$ by
\begin{align}
    J(\pi)
    &= \underset{
    \begin{subarray}{c}
    s_0  \sim p_0(\cdot) \\
    a_t \sim \pi(\cdot|s_t) \\
    s_{t+1} \sim p(\cdot| s_t, a_t) 
    \end{subarray}}{\mathbb{E}} \left [ \sum_{t=0}^\infty \gamma^t R(s_t, a_t) \right ] \, .
\end{align}
An optimal policy $\pi^*$ is one with maximum expected return
\begin{align}
    \pi^* \in \arg \max_\pi J(\pi) \, .
\end{align}

\subsection{Maximum Entropy Reinforcement Learning} \label{sec:intro_maxentrl}
In maximum entropy reinforcement learning (MaxEntRL) an optimal policy $\pi^*$ is approximated by maximizing a surrogate objective function $L(\pi)$, where the reward function from the MDP is extended by an intrinsic reward function. The latter is the (relative) entropy of some particular distribution. A general definition of the MaxEntRL objective function is
\begin{align}
    L(\pi)
    &= \underset{
    \begin{subarray}{c}
    s_0  \sim p_0(\cdot) \\
    a_t \sim \pi(\cdot|s_t) \\
    s_{t+1} \sim p(\cdot| s_t, a_t) 
    \end{subarray}}{\mathbb{E}} \hspace{-8pt} \left [ \sum_{t=0}^\infty \gamma^t \left ( R(s_t, a_t) + \lambda R^{int}(s_t, a_t) \right ) \right ] \, , \label{eq:objective_maxentrl}
\end{align}
where this objective depends on the intrinsic reward function $R^{int}$. We propose a generic formulation that, to the best of our knowledge, encompasses most existing intrinsic rewards from the literature. Given a feature space $\mathcal{Z}$, a conditional feature distribution $q^\pi : \mathcal{S} \times \mathcal{A} \rightarrow \Delta(\mathcal{Z})$, depending on the policy $\pi$, and a relative measure $q^*\in \Delta(\mathcal{Z})$, the MaxEntRL intrinsic reward function is
\begin{align}
    R^{int}(s, a)
    &= - KL_z \left [ q^\pi(z|s, a) \| q^*(z) \right ]
    = \underset{
    \begin{subarray}{c}
    z \sim q^\pi(\cdot|s, a)
    \end{subarray}}{\mathbb{E}} \left [ \log q^*(z) - \log q^\pi(z|s, a) \right ] \label{eq:intr_reward_maxentrl} \, .
\end{align}
Importantly, the intrinsic reward function is (implicitly) dependent on the policy $\pi$ through the distribution $q^\pi$. We define an optimal exploration policy as a policy that maximizes the expected sum of discounted intrinsic rewards only. Note that a policy maximizing $L(\pi)$ is generally not optimal, due to the potential gap between the optimum of the return $J(\pi)$ and the optimum of the learning objective $L(\pi)$. This subject is inherent to exploration with intrinsic rewards \citep{bolland2024behind}.

MaxEntRL algorithms optimize objective functions as defined in equation \eqref{eq:objective_maxentrl} depending on some intrinsic reward function that can be expressed as in equation \eqref{eq:intr_reward_maxentrl}. The particularity of each algorithm is its estimation of the intrinsic reward and of the stochastic gradient of the learning objective. Often, a pseudo reward $\log q^*(z) - \log q^\pi(z|s, a)$ is computed from a sample $z \sim q^\pi(\cdot|s, a)$ to extend the MDP reward function and used by an existing RL algorithm.

Many of the existing MaxEntRL algorithms optimize an objective that depends on the entropy of the policy for exploring the action space \citep{haarnoja2018soft, toussaint2009robot}. The feature space is then the actions space $\mathcal{Z} = \mathcal{A}$, and the conditional feature distribution is the policy $q^\pi(z|s, a) = \pi(z| s)$, for all $a$. Other algorithms optimize objectives enhancing state space exploration \citep{hazan2019provably, lee2019efficient, islam2019marginalized, guo2021geometric}. The feature space is the state space $\mathcal{Z} = \mathcal{S}$. The conditional feature distribution $q^\pi(z|s, a)$ is either the marginal probability of states in trajectories of $T$ time steps, or the discounted state visitation measure, for all $s$ and $a$. In the literature, the relative measure $q^*(z)$ is usually a uniform distribution, and the relative entropy is computed as the differential entropy, i.e., by neglecting $\log q^*(z)$ in equation \eqref{eq:intr_reward_maxentrl}. In continuous spaces, the latter is ill-defined and other relative measures may be used.

\section{MaxEntRL with Visitation Distributions} \label{sec:maxentrl_visitation}

\subsection{Definition of the MaxEntRL Objective} \label{sec:maxentrl_visitation_def}

In the following, we introduce a new MaxEntRL intrinsic reward based on the conditional state-action visitation probability $d^{\pi, \gamma}(\bar s, \bar a| s, a)$ and the conditional state visitation probability $d^{\pi, \gamma}(\bar s| s, a)$
\begin{align}
    d^{\pi, \gamma}(\bar s, \bar a| s, a) &= (1-\gamma) \pi(\bar a| \bar s) \sum_{\Delta=1}^\infty \gamma^{\Delta - 1} p_{\Delta}^\pi(\bar s| s, a) \label{eq:def_conditional_sa_visitation} \\
    d^{\pi, \gamma}(\bar s| s, a) &= (1-\gamma) \sum_{\Delta=1}^\infty \gamma^{\Delta - 1} p_{\Delta}^\pi(\bar s| s, a) \label{eq:def_conditional_s_visitation} \, ,
\end{align}
where $p_\Delta^\pi$ is the transition probability in $\Delta$ time steps with the policy $\pi$. The distribution from equation \eqref{eq:def_conditional_sa_visitation} can be factorized as a function of the distribution from equation \eqref{eq:def_conditional_s_visitation} such that $d^{\pi, \gamma}(\bar s, \bar a| s, a) = \pi(\bar a| \bar s) d^{\pi, \gamma}(\bar s| s, a)$. The conditional state (respectively, state-action) visitation probability distribution measures the future states (respectively, states and actions) that are visited on expectation over infinite trajectories starting from any state and action. Both distributions generalize the (marginal discounted) state visitation probability measure \citep{manne1960linear}.

\begin{definition} \label{def:maxentrl_cv_def} Let us consider the feature space $\mathcal{Z}$ and a feature distribution $h: \mathcal{S} \times \mathcal{A} \rightarrow \Delta(\mathcal{Z})$. The intrinsic reward is defined by equation \eqref{eq:intr_reward_maxentrl}, for any relative measure $q^*$, with conditional distribution
\begin{align}
    q^\pi(z| s, a) &= \int h(z| \bar s, \bar a) d^{\pi, \gamma}(\bar s, \bar a| s, a) \: d\bar s \: d\bar a \, . \label{eq:def_q}
\end{align}
\end{definition}

Optimal exploration policies are here intrinsically motivated to take actions so that the discounted visitation measure of future features is distributed according to $q^*$ in each state and for each action. It allows to select features that must be visited during trajectories according to prior knowledge about the problem if any. Alternatively, it allows to only explore lower dimensional feature spaces, or to explore sufficient statistics from the state-action pairs.

The MaxEntRL objective from Definition \ref{def:maxentrl_cv_def} can be optimized by any existing RL algorithm that is adapted to compute for each state $s$ and action $a$ the additional (pseudo) reward
\begin{align}
    R^{int}(s, a) &= \log q^*(z) - \log q^\pi(z| s, a) \, , \label{eq:def_approx_r}
\end{align}
where $z \sim q^\pi(\cdot | s, a)$. This reward is a single-sample Monte-Carlo estimate of equation \eqref{eq:intr_reward_maxentrl}, unbiased for fixed $q^\pi$. This computation requires sampling features $z$ from the conditional distribution $q^\pi$ and estimating the probability of these samples $q^\pi(z| s, a)$. It can be achieved by solving the integral equation \eqref{eq:def_q}, e.g., numerically by sampling states $\bar s \sim d^{\pi, \gamma}(\cdot| s, a)$, actions $\bar a \sim \pi(\cdot| \bar s)$, and finally features $z \sim h(\cdot| \bar s, \bar a)$. This particular sampling procedure requires access to the unknown conditional state visitation probability. Section \ref{sec:off_pol_algo} provides a method for learning that distribution off-policy.

\subsection{Relationship with Alternative MaxEntRL Objectives}

Let us relate MaxEntRL with the new intrinsic reward function to the maximization of a lower bound on the state-action value function of the MDP (computed without intrinsic rewards). We rely on Theorem \ref{thr:q_bound}, shown in Appendix \ref{apx:proofs}, close to the results from \cite{kakade2002approximately}.

\begin{theorem} \label{thr:q_bound} Let the reward function $R(s, a)$ be non-negative, let $\pi$ be a policy with state-action value function $Q^{\pi}(s, a)$, and let the visitation measures be non-zero over their support, then,
\begin{align}
    Q^{\pi}(s, a)
    &\geq Q^{\pi^*}(s, a)\, \exp \left ( - \left \| \log \frac{d^{\pi, \gamma}(\cdot, \cdot| s, a)}{d^{\pi^*, \gamma}(\cdot, \cdot| s, a)} \right \|_{\infty} \right ) \, , \label{eq:bound_thr_1}
\end{align}
where $\| f \|_\infty = \sup_x | f(x)|$ is the $L_\infty$-norm of $f$.
\end{theorem}

Let us again consider that the feature distribution $h$ is the identity map, so that $z = (\bar s, \bar a)$, and apply the triangle inequality on equation \eqref{eq:bound_thr_1}. For any policy $\pi$, we get the bound
\begin{align}
    Q^{\pi}(s, a)
    &\geq Q^{\pi^*}(s, a)\, \exp \left ( - \left \| \log \frac{d^{\pi, \gamma}(\cdot, \cdot| s, a)}{q^*(\cdot, \cdot)} \right \|_{\infty} \right ) \exp \left ( - \left \| \log \frac{d^{\pi^*, \gamma}(\cdot, \cdot| s, a)}{q^*(\cdot, \cdot)} \right \|_{\infty} \right ) \, . \label{eq:bound_q}
\end{align}
The bound on the state-action value function of any policy $\pi$ in equation \eqref{eq:bound_q} is an exponentially decreasing function of the two error terms $\| \log d^{\pi, \gamma}(\cdot, \cdot| s, a) - \log q^*(\cdot, \cdot) ||_{\infty}$ and $\| \log d^{\pi^*, \gamma}(\cdot, \cdot| s, a) - \log q^*(\cdot, \cdot) ||_{\infty}$. The first can be minimized as a function of $\pi$ while the second is independent of the policy, and can thus not be reduced. Let us assume that an optimal exploration policy has zero expected discounted sum of intrinsic rewards, and that the target measure and the visitation measures are smooth. Then, an optimal exploration policy maximizes the bound in equation \eqref{eq:bound_q}. Optimizing the MaxEntRL objective we introduce can be seen as a practical algorithm to compute a policy that maximizes the lower bound equation \eqref{eq:bound_q}. The quality of the resulting policy then only depends on the choice of the distribution $q^*$.

\section{Off-policy Learning of Conditional Visitation Models} \label{sec:off_pol_algo}

\subsection{Fixed-Point Properties of Conditional Visitation}

As explained in Section \ref{sec:maxentrl_visitation_def}, the MaxEntRL intrinsic reward function in Definition \ref{def:maxentrl_cv_def} can be computed from samples of the conditional distribution $q^\pi(z|s, a)$, which in turn can be computed based on samples of the conditional state visitation distribution $d^{\pi, \gamma}(\bar s|s, a)$. In this section, we establish useful properties of this visitation distribution.

Let us first recall that the conditional state visitation distribution accepts a recursive definition \citep{janner2020generative} that is a trivial fixed point of the operator $\mathcal{T}^\pi$ from Definition \ref{def:operator_T}.

\begin{definition} \label{def:operator_T} The operator $\mathcal{T}^\pi$ is defined over the space of conditional state distribution as
\begin{align}
    \mathcal{T}^\pi q(\bar s| s, a) 
    &= (1-\gamma) p(\bar s| s, a) + \gamma
    \underset{
        \begin{subarray}{c}
            s' \sim p(\cdot| s, a) \\
            a' \sim \pi(\cdot | s')
        \end{subarray}}{\mathbb{E}}  \left [ q(\bar s| s', a')  \right ] \, .
\end{align}
\end{definition}

In Theorem \ref{thr:contraction_T}, we establish that the operator $\mathcal{T}^\pi$ is a contraction mapping, which furthermore implies the uniqueness of its fixed point. Assuming the result of the operator could be computed (or estimated), the fixed point could also be computed by successive application of this operator. It would allow computing the conditional state visitation distribution and the intrinsic reward function.

\begin{theorem} \label{thr:contraction_T} The operator $\mathcal{T}^\pi$ is $\gamma$-contractive in $\bar L_n$-norm, where $\bar L_n( f )^n = \sup_y \int | f(x| y) |^n \: dx$.
\end{theorem}
The theorem is shown in Appendix \ref{apx:proofs}.

\subsection{TD Learning of Conditional Visitation Models} \label{sec:learning_visitation}
In practice, computing the result of the operator $\mathcal{T}^\pi$ (and $(\mathcal{T}^\pi)^N$ after $N$ applications) may be intractable when large or continuous state-action spaces are at hand. It furthermore requires having a model of the MDP. A common alternative approach is to rely on a function approximator $d_\psi$ to approximate the fixed point. Theorem \ref{thr:contraction_T} suggests optimizing the parameters of this model $d_\psi$ to minimize the residual of the operator, measured with the $\bar L_n$-norm for which the operator is $\gamma$-contractive, similarly to TD-learning methods \citep{sutton2018reinforcement}. Nevertheless, measuring the residual with the $\bar L_n$-norm requires estimating the MDP transition function \citep{janner2020generative}, and can therefore not be trivially minimized by stochastic gradient descent using sampled transitions. We therefore propose to solve as surrogate a minimum cross-entropy problem, in which stochastic gradient descent can be applied afterward. For any policy $\pi$, the distribution is approximated with a function approximator $d_\psi$ with parameter $\psi$ optimized to solve
\begin{align}
    \arg \min_\psi
    \underset{
        \begin{subarray}{c}
            s, a \sim g(\cdot, \cdot) \\
            \bar s \sim \left (\mathcal{T}^\pi \right )^N d_\psi(\cdot| s, a)
        \end{subarray}}{\mathbb{E}} \left [ - \log d_\psi(\bar s| s, a) \right ] \, , \label{eq:mle_optim_visitation_model}
\end{align}
where $g$ is an arbitrary distribution over the state and action spaces, and where $N$ is any positive integer. This optimization problem is related to minimizing the KL-divergence instead of an $\bar L_n$-norm \citep{bishop2006pattern}.

Let us make explicit how samples from the distribution $(\mathcal{T}^\pi )^N d_\psi(\bar s| s, a)$ can be generated from the MDP. By definition of the operator $\mathcal{T}^\pi$, the distribution $(\mathcal{T}^\pi )^N d_\psi(\bar s| s, a)$ is the mixture
\begin{align}
    (\mathcal{T}^\pi )^N d_\psi(\bar s| s, a)
    &= \left ( \sum_{\Delta=1}^{N} (1 - \gamma) \gamma^{\Delta - 1} p_\Delta^\pi(\bar s| s, a) \right ) + \gamma^N \underset{
        \begin{subarray}{c}
            s' \sim p_N^\pi(\cdot| s, a) \\
            a' \sim \pi(\cdot |s')
        \end{subarray}}{\mathbb{E}} \left [ d_\psi(\bar s| s', a') \right ] \\
	&= \sum_{\Delta = 1}^\infty \mathcal{G}_{1-\gamma}(\Delta) b^\beta_{\psi, \pi}(\bar s|s, a, \Delta) \big |_{\beta = \pi} \, ,
\end{align}
where $\mathcal{G}_{1-\gamma}(\Delta)$ is the probability of $\Delta$ from a geometric distribution of parameter $1-\gamma$, and
\begin{align}
	b^\beta_{\psi, \pi}(\bar s|s, a, \Delta) = &
	\left\{ 
		\begin{array}{ll}
    	p_\Delta^\beta(\bar s| s, a) & \Delta \leq N \\[0.5em]
    	\underset{
        \begin{subarray}{c}
            s' \sim p_N^\beta(\cdot| s, a) \\
            a' \sim \pi(\cdot |s')
        \end{subarray}}{\mathbb{E}} \left[ d_\psi(\bar s| s', a') \right] & \Delta > N
  		\end{array}
	\right. \, .
\end{align}
Sampling from $(\mathcal{T}^\pi )^N d_\psi(\bar s| s, a)$ consists in sampling from the mixture. First, $\Delta$ is drawn from a geometric distribution of parameter $1-\gamma$. Second, a state is sampled as $\bar s \sim p_\Delta^\pi(\cdot| s, a)$ if $\Delta \leq N$ or as $\bar s \sim d_\psi(\cdot| s', a')$ otherwise; where $s' \sim p_N^\pi(\cdot| s, a)$ and $a' \sim \pi(\cdot| s')$.

Let us reformulate the problem equation \eqref{eq:mle_optim_visitation_model} to highlight the previous sampling procedure, and such that it can be estimated from transitions sampled from an arbitrary policy $\beta$ in the MDP. To that end, we apply importance weighting and get the equivalent optimization problem
\begin{align}
	\arg \min_\psi
    &\underset{
        \begin{subarray}{c}
            s, a \sim g(\cdot, \cdot) \\
            \Delta \sim \mathcal{G}_{1-\gamma}(\cdot) \\
            \bar s \sim b^\beta_{\psi, \pi}(\cdot|s, a, \Delta)
        \end{subarray}}{\mathbb{E}} \left [ - \frac{b^\pi_{\psi, \pi}(\bar s|s, a, \Delta)}{b^\beta_{\psi, \pi}(\bar s|s, a, \Delta)} \log d_\psi(\bar s| s, a) \right ] \, . \label{eq:mle_optim_visitation_model_v2}
\end{align}
In the particular cases where $\beta=\pi$ or where $N=1$, the importance weight simplifies to one, otherwise it can be simplified to a (finite) product of ratios of policies. 

Learning $d_\psi$ from samples can be achieved by solving problem equation \eqref{eq:mle_optim_visitation_model_v2} as an intermediate step to any RL algorithm. First, the objective function is estimated as described using transitions stored in a batch or generated with a behavior policy $\beta$. The sample $\bar s = s_{t + \Delta}$ is available in the batch or replay buffer if $\Delta \leq N$, or $\bar s \sim d_{\psi}(\cdot| s_{t+N}, a_{t+N}')$ is bootstrapped otherwise; where $a_{t+N}' \sim \pi(\cdot| s_{t+N})$ and $\Delta \sim \mathcal{G}_{1-\gamma}(\cdot)$. Second, this estimate is differentiated, and the parameter $\psi$ is updated by gradient descent steps. In practice, the gradients generated by differentiating this loss function are biased. The influence of the parameter $\psi$ on the probability of the sample $z$ is neglected when bootstrapping, i.e., the partial derivative of $\left (\mathcal{T}^\pi \right )^N d_\psi(\bar s| s_t, a_t)$ with respect to $\psi$ is neglected, and a target network is used. This is analogous to SARSA and TD-learning strategies \citep{sutton2018reinforcement}. Furthermore, we suggest neglecting the importance weights, which introduces a dependency of the distribution $d_\psi$ on the policy $\beta$. Finally, the model $d_\psi$ is used to compute the intrinsic rewards and update the policy.

\section{Experiments} \label{sec:experiments}

\subsection{Experimental Setting}

Illustrative experiments are performed on adapted environments from the Minigrid suite \citep{MinigridMiniworld23}. In the latter, an agent must travel across a grid containing walls and passages in order to reach a goal. The size of the grid and the number of passages and walls depend on the environment. The state space is composed of the agent's orientation, its position on the grid, as well as the positions of the passages in the walls and their orientations. In some environments, the goal to be reached is randomly generated and is also part of the state. The agent can take four different actions: turn left, turn right, move forward, or stand still. The need for exploration comes from the sparsity of the reward function, which is zero everywhere and equals one in the state to be reached.

As explained, the model $d_\psi$ is learned during an intermediate step added to an arbitrary RL algorithm that evaluates and optimizes the intrinsic rewards with the MDP rewards. Experiments were performed using off-policy actor-critic \citep{degris2012off}, i.e., an approximate policy iteration algorithm, adapted as advocated. This new algorithm is detailed in Appendix \ref{apx:algo} and is called off-policy actor-critic with conditional visitation measures (OPAC+CV) in the remainder of the paper. For the Minigrid environments, the features $z \in \mathcal{Z}$ are the pairs of horizontal and vertical positions of the agent in the environment, the function $h$ is a deterministic mapping that computes these positions based on the state-action pairs, and the relative measure $q^*$ is uniform. The pseudo-code is provided in Appendix \ref{apx:algo}, and the implementation choices and hyperparameters are detailed in Appendix \ref{apx:parameters}.

This new MaxEntRL algorithm is compared to two alternative algorithms. The first concurrent method is soft actor-critic (SAC) \citep{haarnoja2018soft}. It is a commonly-used MaxEntRL algorithm where the feature space is the action space $\mathcal{Z} = \mathcal{A}$, the conditional distribution is the policy $q^\pi(z|s, a) = \pi(z| s)$ for all $a$, and the relative measure $q^*$ is uniform. To the best of our knowledge, the MaxEntRL objective optimized in soft actor-critic is also the only alternative where policies can be computed off-policy when the state and action space is large or continuous.

The second concurrent method intrinsically motivates agents to have uniform (marginal) discounted visitation measures as originally proposed by \citet{hazan2019provably} and discussed in Section \ref{sec:introduction}. To that end, we adapt the algorithm from \citet{zhang2021exploration}. First, to improve sample efficiency, policies are optimized using off-policy actor-critic \citep{degris2012off} instead of PPO \citep{schulman2017proximal}. Second, we use a categorical distribution rather than a variational auto-encoder to approximate the visitation measure, which is made possible as the state-action space is discrete. It allows optimizing the approximator without relying on the evidence lower bound. We refer to that adapted algorithm as off-policy actor-critic with marginal visitation measures (OPAC+MV). Here, the feature space $\mathcal{Z}$ is the same as in OPAC+CV, the conditional distribution $q^\pi(z|s, a)$ is the discounted visitation measure of features for each state $s$ and action $a$, and the relative measure $q^*$ is uniform. In practice, the feature probability and intrinsic reward are computed as for OPAC+CV; more details are available in Appendix \ref{apx:algo}. Even if off-policy actor-critic is off-policy, learning the model of the visitation measure requires online buffer updates. The final algorithm is therefore on-policy.

\subsection{Exploring Sparse-Reward Environments}

\begin{figure}[b]
  \centering
  \includegraphics[width=0.8\linewidth]{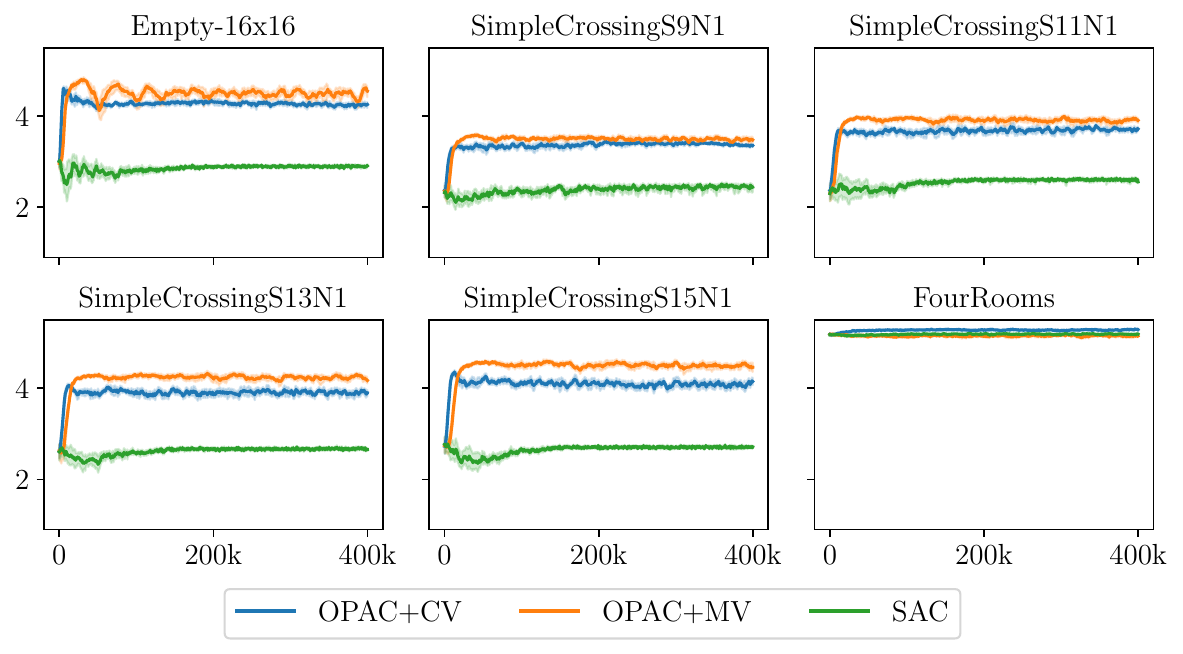}
  \caption{Evolution of the entropy of the discounted visitation probability measure of the position of the agent on the grid when computing exploration policies (i.e., when neglecting the rewards of the MDP). The entropy is computed empirically with Monte Carlo simulations. For each iteration, the interquartile mean over 15 runs is reported, along with its $95\%$ confidence interval.
  }
  \label{fig:learning_curve_entropy_exploration}
\end{figure}

\begin{figure}[t]
  \centering
  \includegraphics[width=0.8\linewidth]{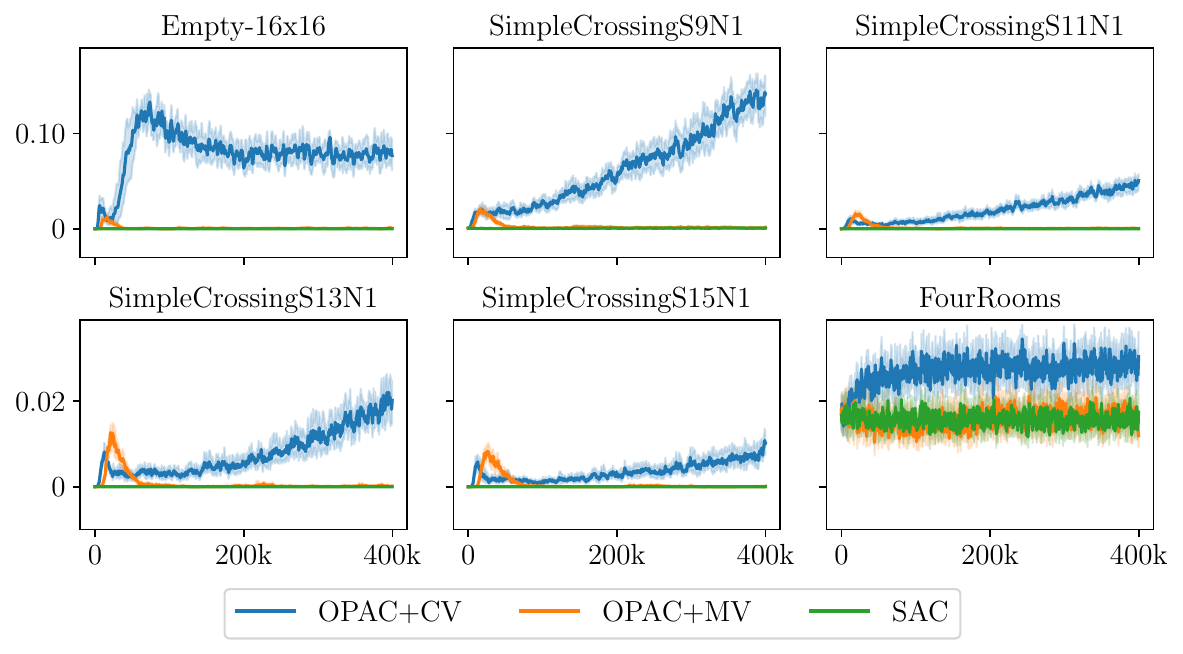}
  \caption{Expected return during the (exploration) policy optimization with OPAC+CV and OPAC+MV. The expectation is computed empirically with Monte Carlo simulations. For each iteration, the interquartile mean over 15 runs is reported, along with its $95\%$ confidence interval.}
  \label{fig:learning_curve_return_exploration}
\end{figure}

The feature space coverage of optimal exploration policies computed with OPAC+CV, OPAC+MV, and SAC is first compared. In Figure \ref{fig:learning_curve_entropy_exploration}, the evolution of the entropy of the discounted visitation measure of features is shown as a function of the number of algorithm iterations, when only the intrinsic rewards are considered. For each environment, the entropy increases rapidly with the OPAC+CV and OPAC+MV algorithms, and a high-entropy policy results from the optimization. In most environments, OPAC+MV achieves the highest entropy, followed closely by OPAC+CV, while SAC performs poorly. It is worth noting that OPAC+CV performs competitively with concurrent method despite optimizing a different objective than the reported discounted visitation measure.

In Figure \ref{fig:learning_curve_return_exploration}, the evolution of the expected returns of the policies is reported during learning. As can be seen, optimizing the exploration objective presented in Section \ref{sec:maxentrl_visitation_def} with OPAC+CV provides optimal exploration policies with significantly higher expected return compared to OPAC+MV and SAC. Importantly, comparing Figure \ref{fig:learning_curve_entropy_exploration} and Figure \ref{fig:learning_curve_return_exploration}, one can see that policies with small differences in the entropy of the discounted visitation measure may achieve very different expected returns.

In the literature, feature exploration is usually used to compute optimal exploration policies as an initialization when extrinsic rewards are not available. Our method is an off-policy alternative yielding policies with good feature space coverage and larger expected return.

\subsection{Controlling Sparse-Reward Environments}

\begin{figure}[t]
  \centering
  \includegraphics[width=0.8\linewidth]{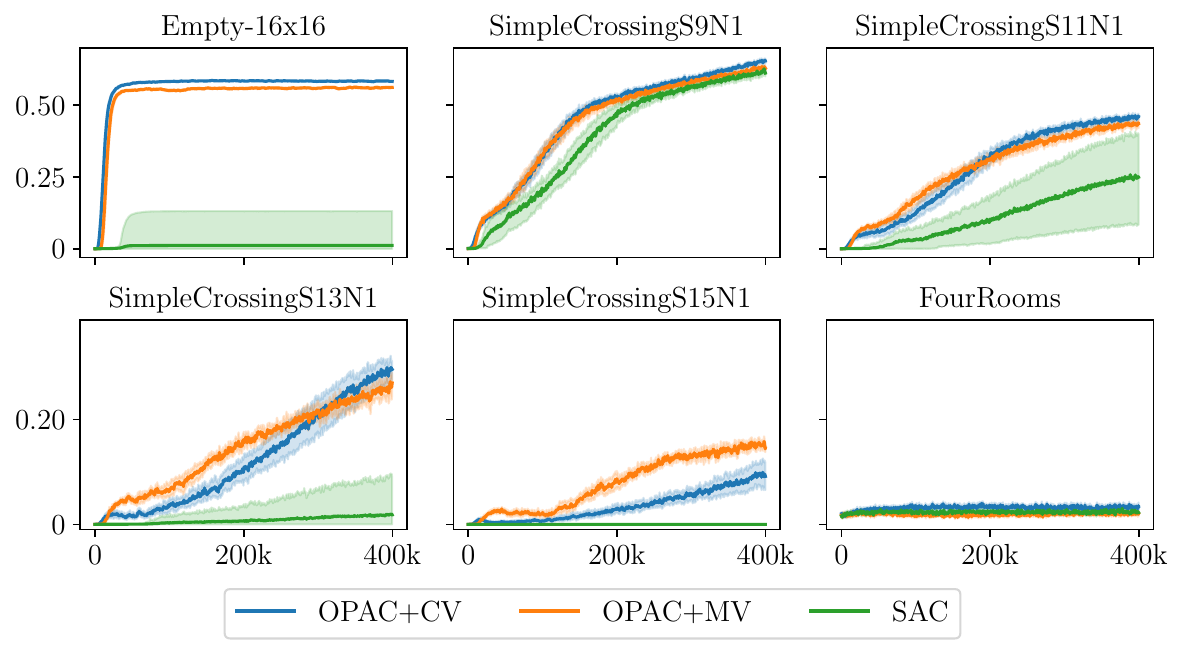}
  \caption{Expected return during the policy optimization with OPAC+CV, OPAC+MV, and SAC. The expectation is computed empirically with Monte Carlo simulations. For each iteration, the interquartile mean over 15 runs is reported, along with its $95\%$ confidence interval.}
  \label{fig:learning_curve_return_control}
\end{figure}

The objective of MaxEntRL is to provide intrinsic motivations to explore in order to compute a high-performance policy. In Figure \ref{fig:learning_curve_return_control}, the expected return of OPAC+CV is compared to that of SAC and OPAC+MV. As can be seen, our method always performs at least as well as SAC. In the \texttt{SimpleCrossing}-environments, the two methods perform equivalently for the first one, OPAC+CV performs similarly to the lucky realizations of SAC for the second one, and only OPAC+CV computes (with high probability) policies with non-zero return for the last two. These environments are open grids of different sizes where the agent shall cross a wall through a small passage to reach the target. The larger the environment, the lower the probability of reaching the goal with a uniform policy, and the worse the performance of SAC. The same can be observed in the \texttt{Empty-16x16}-environment. On the contrary, both MaxEntRL methods perform equivalently in the \texttt{FourRooms}-environment, where complex exploration is apparently not necessary to solve the problem. Finally, our method slightly outperforms OPAC+MV in all environments, except in \texttt{SimpleCrossingS15N1} where the concurrent method performs best. Two factors may influence the performance. First, the intrinsic reward functions have different scales, and the weight $\lambda$ is constant. Second, the expected returns of optimal exploration policies are different; see Figure \ref{fig:learning_curve_return_exploration}. Probably the most important is that both methods allow computing policies with non-zero rewards. With an appropriate scheduling on $\lambda$, both methods could eventually compute high-performing policies.

\subsection{Discussion of Experiments}
Several phenomena influence the learning of the visitation model. First, when $\gamma$ is close to one, the learning becomes unstable in practice. We hypothesize this results from the increased importance of future states. Increasing parameter $N$ helps mitigate the issue as there is less bootstrapping, reducing the risk of learning a biased target. Second, we neglect the importance weights in practice to reduce variance, which makes $d_\psi$ partially dependent on the behavior policy $\beta$. In our online setting, a relatively small buffer is refreshed sufficiently often to mitigate this dependence. Bootstrapping still propagates long-term effects of the policy, which also allows batch-mode RL with a biased model. 

In the experiments, OPAC+CV reached strong marginal coverage quickly. In general, our exploration objective is not expected to improve the marginal visitation coverage compared to its direct maximization. In practice, the off-policiness and the stability noted above can still make OPAC+CV competitive. We also observed that maximizing our objective leads to higher conditional visitation entropy, given the initial state, meaning a wider set of features is explored within each independent trajectory. This could explain the higher returns observed in Figure \ref{fig:learning_curve_return_exploration}.

Finally, we relied on off-policy actor-critic for concreteness, yet the MaxEntRL objective is agnostic to the control backbone, and similar results should hold with other RL methods. Our method offers a practical alternative to directly maximizing marginal visitation, without focusing on the potential theoretical advantages of different exploration objectives.

\section{Conclusion} \label{sec:conclusion}

In this paper, we presented a new MaxEntRL objective providing intrinsic reward bonuses proportional to the entropy of the distribution of features built from the states and actions visited by the agent in future time steps. The reward bonus can be estimated efficiently by sampling from the conditional distribution of states visited, which we proved to be the fixed point of a contraction mapping. It can therefore be learned for any policy relying on batches of arbitrary transitions. We proposed an end-to-end off-policy algorithm maximizing our objective that allows exploring effectively the state and action spaces. The algorithm is benchmarked on several control problems. The method we developed is easy to implement and can be integrated into already existing RL algorithms.

In this paper, experiments were limited to relatively small-scale environments. Future work should focus on benchmarking the method in more challenging environments, including environments with larger or continuous state-action spaces. For the continuous case, this will require adapting the density estimator and the algorithm accordingly. Furthermore, in this paper, the feature space to explore is fixed a priori, but could be learned. A potential avenue is to explore reward-predictive feature spaces. Finally, the distribution that is learned for exploration purposes can be used to generate new samples to enhance sample efficiency when learning the critic. The integration of this approach into the MaxEntRL framework is left for future work.

\bibliography{bibliography}
\bibliographystyle{apalike}
\setcitestyle{authoryear,round,citesep={;},aysep={,},yysep={;}}

\newpage
\appendix

\section{Proofs of Theorems} \label{apx:proofs}

\paragraph{Proof Theorem \ref{thr:q_bound}.} Let us express the state-action value function as a function of the conditional state-action visitation distribution \citep{eysenbach2020c, janner2020generative}
\begin{align}
    Q^\pi(s, a) 
    &= \frac{1}{1-\gamma} \int d^{\pi, \gamma}(\bar s, \bar a| s, a) R(\bar s, \bar a) \: d\bar s \: d\bar a \nonumber \\
    &= \frac{1}{1-\gamma} \int \frac{d^{\pi, \gamma}(\bar s, \bar a| s, a)}{d^{\pi^*, \gamma}(\bar s, \bar a| s, a)} d^{\pi^*, \gamma}(\bar s, \bar a| s, a) R(\bar s, \bar a) \: d\bar s \: d\bar a \nonumber \\
    &\geq Q^{\pi*}(s, a) \inf_{\bar s, \bar a} \frac{d^{\pi, \gamma}(\bar s, \bar a| s, a)}{d^{\pi^*, \gamma}(\bar s, \bar a| s, a)} \label{eq:monotone_ineq} \\
    &= Q^{\pi*}(s, a) \exp \inf_{\bar s, \bar a} \left ( \log \frac{d^{\pi, \gamma}(\bar s, \bar a| s, a)}{d^{\pi^*, \gamma}(\bar s, \bar a| s, a)} \right ) \nonumber \\
    &= Q^{\pi*}(s, a) \exp \left ( \inf_{\bar s, \bar a} \left (  \log d^{\pi, \gamma}(\bar s, \bar a| s, a) - \log d^{\pi^*, \gamma}(\bar s, \bar a| s, a) \right ) \right ) \nonumber \\
    &= Q^{\pi*}(s, a) \exp \left ( -\sup_{\bar s, \bar a} \left ( \log d^{\pi^*, \gamma}(\bar s, \bar a| s, a) - \log d^{\pi, \gamma}(\bar s, \bar a| s, a) \right ) \right ) \nonumber \\
    &\geq Q^{\pi*}(s, a) \exp \left ( -\sup_{\bar s, \bar a} \left | \log d^{\pi^*, \gamma}(\bar s, \bar a| s, a) - \log d^{\pi, \gamma}(\bar s, \bar a| s, a) \right | \right ) \label{eq:sup_of_abs} \\
    &= Q^{\pi*}(s, a) \exp \left ( - \|\log d^{\pi^*, \gamma}(\cdot, \cdot| s, a) - \log d^{\pi, \gamma}(\cdot, \cdot| s, a)\|_\infty \right ) \nonumber \, .
\end{align}
Inequation \eqref{eq:monotone_ineq} holds by the monotonicity of the (Lebesgue) integral, and inequation \eqref{eq:sup_of_abs} holds as $\sup_x f(x) \leq \sup_x | f(x) |$ for any function $f$.

\hfill $\square$

\paragraph{Proof Theorem \ref{thr:contraction_T}.} For all conditional distributions $q$ and $q'$
\begin{align*}
    \bar L_n(\mathcal{T}^\pi q(\cdot| s, a), \mathcal{T}^\pi q'(\cdot| s, a))^n &= \sup_{s, a} \int | \mathcal{T}^\pi q(\bar s| s, a) - \mathcal{T}^\pi q'(\bar s| s, a) |^n \: d \bar s \\
    &= \gamma \sup_{s, a} \int \left |
    \underset{
        \begin{subarray}{c}
            s' \sim p(\cdot| s, a) \\
            a' \sim \pi(\cdot | s')
        \end{subarray}}{\mathbb{E}}  \left [ q(\bar s| s', a') - q'(\bar s| s', a') \right ]  \right |^n \: d \bar s \\
    &\leq \gamma \sup_{s, a} \int
    \underset{
        \begin{subarray}{c}
            s' \sim p(\cdot| s, a) \\
            a' \sim \pi(\cdot | s')
        \end{subarray}}{\mathbb{E}}  \left [  | q(\bar s| s', a') - q'(\bar s| s', a') |^n \right ] \: d \bar s \\
    &= \gamma \sup_{s, a}
    \underset{
        \begin{subarray}{c}
            s' \sim p(\cdot| s, a) \\
            a' \sim \pi(\cdot | s')
        \end{subarray}}{\mathbb{E}}  \left [ \int | q(\bar s| s', a') - q'(\bar s| s', a') |^n\: d \bar s \right ] \\
    &\leq \gamma \sup_{s, a} \sup_{s', a'} \left (  \int | q(\bar s| s', a') - q'(\bar s| s', a') |^n \: d \bar s \right ) \\
    &= \gamma \sup_{s', a'}  \int | q(\bar s| s', a') - q'(\bar s| s', a') |^n \: d \bar s \\
    &= \gamma \bar  L_n(q(\cdot| s, a), q'(\cdot| s, a))^n \, .
\end{align*}

\hfill $\square$

\newpage
\section{Off-Policy RL with Conditional Visitation Measure} \label{apx:algo}

In the following, we adapt soft actor-critic \citep{haarnoja2018soft}, itself an adaptation of off-policy actor-critic \citep{degris2012off}, according to the procedure from Section \ref{sec:off_pol_algo}. In essence, soft actor-critic estimates the state-action value function with a parameterized critic $Q_\phi$, which is learned using expected SARSA (sometimes called generalized SARSA), and updates the parameterized policy $\pi_\theta$ with approximate policy iteration (i.e., off-policy policy gradient), all based on one-step transitions stored in a replay buffer $\mathcal{D}$. The actor and critic loss functions are furthermore extended with the log-likelihood of actions weighted by the parameter $\lambda_{SAC}$, therefore called soft and considered a MaxEntRL algorithm using the entropy of policies as intrinsic reward. In the particular case where $\lambda$ equals zero, the algorithm boils down to a slightly revisited implementation of off-policy actor-critic.

Soft actor-critic is adapted to MaxEntRL with the intrinsic reward function defined in Section \ref{sec:maxentrl_visitation_def}, as follows. First, $N$-step transitions are stored in the buffer $\mathcal{D}$ instead of one-step transitions. Second, the conditional state visitation distribution is estimated with a function approximator $d_\psi$ and learned with stochastic gradient descent. Third, at each iteration of the critic updates, the reward provided by the MDP is extended with the intrinsic reward.

Formally, the parameterized critic $Q_\phi$ is iteratively updated by performing stochastic gradient descent steps on the loss function
\begin{align}
    \mathcal{L}(\phi) &=
    \underset{
        \begin{subarray}{c}
            s_t, a_t \sim \mathcal{D}
        \end{subarray}}{\mathbb{E}}  \left [ \left ( Q_\phi(s_t, a_t) - y \right )^2 \right ] \label{eq:loss_critic} \\
	y &= R(s_t, a_t) + \lambda R^{int}(s_t, a_t) + \gamma \left ( Q_{\phi'}(s_{t+1}, a_{t+1'}) - \lambda_{SAC} \log \pi_\theta(a_{t+1'}| s_{t+1}) \right ) \, ,
\end{align}
where $a_{t+1'} \sim \pi_\theta(\cdot| s_{t+1})$, and where $\phi'$ is the target network parameter. 

Furthermore, the policy $\pi_\theta$ is updated performing gradient descent steps on the loss function
\begin{align}
    \mathcal{L}(\theta) &= -
    \underset{
        \begin{subarray}{c}
            s_t, a_t \sim \mathcal{D}
        \end{subarray}}{\mathbb{E}}  \left [ \log \pi_\theta(a_{t'}| s_t) A(s_t, a_{t'}) \right ] \label{eq:loss_actor}\\
        A(s_t, a_{t'}) &= Q_\phi(s_t, a_{t'}) - \lambda_{SAC} \log \pi_\theta(a_{t'}| s_t) \, ,
\end{align}
where $a_{t'} \sim \pi_\theta(\cdot| s_t)$.

Algorithm \ref{algo:offpol_exploration} summarizes the learning steps during each iteration.\footnote{\scriptsize Implementation: \url{https://github.com/adrienBolland/future-visitation-exploration}} It differs slightly from the original soft actor-critic \citep{haarnoja2018soft}. The loss equation \eqref{eq:loss_actor} is based on the log-trick instead of the reparametrization trick, the expected SARSA update in equation \eqref{eq:loss_critic} is approximated by sampling, and a single value function is learned, as implemented in CleanRL \citep{huang2022cleanrl}. These changes are of minor importance in our experiments.

\begin{algorithm}
\caption{SAC with conditional visitation measure for exploration}
\label{algo:offpol_exploration}
\begin{algorithmic}
\STATE Initialize the policy $\pi_\theta$, the soft critic $Q_\phi$, and the visitation model $d_\psi$
\STATE Initialize the critic target $Q_{\phi'}$ and visitation target $d_{\psi'}$
\STATE Initialize the replay buffer with random $N$-step transitions
\WHILE{Learning}
    \STATE Sample transitions from the policy $\pi_\theta$ and add them to the buffer
    \WHILE{Update the visitation model}
    	\STATE Sample a batch of $N$-step transitions from the buffer
        \STATE Perform a stochastic gradient descent step on $\mathcal{L}(\psi)$
    \ENDWHILE
    \WHILE{Update the critic}
    	\STATE Sample a batch of $N$-step transitions from the buffer (use only the 1-step transitions)
    	\STATE For each element of the batch sample $z_t \sim q^\pi(\cdot| s_t, a_t)$
    	\STATE Estimate the intrinsic reward $R^{int}(s_t, a_t) = \log q^*(z_t) - \log q^\pi(z_t|s_t, a_t)$
    	\STATE Perform a stochastic gradient descent step on $\mathcal{L}(\phi)$
    \ENDWHILE
    \STATE Sample a batch of $N$-step transitions from the buffer (use only the 1-step transitions)
    \STATE Perform a stochastic gradient descent step on $\mathcal{L}(\theta)$
    \STATE Update the target parameters with Polyak averaging
\ENDWHILE
\end{algorithmic}
\end{algorithm}

\section{Hyperparameters Experiments} \label{apx:parameters}

In practice, the agent observes the concatenation of the one-hot-encoding of the components of the state space and takes actions in one-hot-encoding format too. The policy $\pi_\theta$ is a neural network that outputs a categorical distribution over the action representation. The critic $Q_\phi$ is a neural network that takes as input the concatenation of the state and action representations and outputs a scalar. In OPAC+CV, the visitation distribution model $d_\psi$ is also a neural network that takes the same input as the critic $Q_\phi$ and outputs, for each component of the state space, a categorical distribution over its one-hot-encoding representation. In OPAC+MV, the visitation distribution model $d_\psi$ is a marginal distribution over the same one-hot-encoding representation. In both algorithms, this amounts to assuming the conditional independence of the future state components given the state and action taken as input. This implementation choice mitigates the curse of dimensionality. In addition, it allows computing the probability of a feature in closed form. The probability equals the product of the probability of the vertical position and the probability of the horizontal position provided in one-hot-encoding by the model $d_\psi$. Table \ref{tab:parameters} summarizes the hyperparameters used in the experiments. In practice, the parameter $\lambda_{SAC}$ is constant for SAC, OPAC+CV, and OPAC+MV simulations.

\begin{table}[p]
\centering
\caption{Hyperparameters}
\label{tab:parameters}
\begin{tabular}{l|c}
	Parameter & Value \\
	\hline
	Neurons for each network layers & $256$ \\
	Layers policy & $2$ \\
	Layers critic & $2$  \\
	Learning rate policy & $10^{-5}$ \\
	Learning rate critic & $10^{-4}$ \\
	Maximum trajectory length & $200$ \\
	Buffer size & $1000$ \\
	Batch size & $32$ \\
	Critic target update weight $\tau$ & $0.1$ \\
	Discount factor $\gamma$ & $0.98$ \\
	\hline
	SAC $\lambda_{SAC}$ & $0.002$ \\
	\hline
	Layers visitation model OPAC+CV & $2$ \\
	Learning rate visitation model & $10^{-5}$ \\
	MaxEntRL $\lambda$ & $0.01$ \\
	Density model target update weight $\tau$ & $1$ \\
	Bootstrap horizon $N$ & $10$
\end{tabular}
\end{table}

\end{document}